\documentclass{frontiersinFPHY_FAMS}

\makeatletter
\@twocolumnfalse 
\def\@extraAuth{}
\makeatother

\newcommand{\Address}{} 

\usepackage{url,hyperref,lineno,microtype}
\usepackage[onehalfspacing]{setspace}
\usepackage{booktabs}
\usepackage{siunitx}
\usepackage{amsmath}
\usepackage{amssymb}

\def\firstAuthorLast{Beirami {et~al.}}

\title[VA Dice Loss for HNC]{Improving Deep Learning-Based Target Volume Auto-Delineation for Adaptive MR-Guided Radiotherapy in Head and Neck Cancer: Impact of a Volume-Aware Dice Loss} 

\author[\firstAuthorLast]{Sogand Beirami$^{1}$, Zahra Esmaeilzadeh$^{1}$, Ahmed Gomaa$^{1}$, Pluvio Stephan$^{1}$, Ishita Sheth$^{1}$, Thomas Weissmann$^{1}$, Juliane Szkitsak$^{1}$, Philipp Schubert$^{1}$, Yixing Huang$^{1}$, Annette Schwarz$^{1}$, Stefanie Corradini$^{1}$, and Florian Putz$^{1,*}$}

\address{$^{1}$Strahlenklinik, Universitätsklinikum Erlangen, Germany}

\correspondance{}
\def\corrAuthor{PD Dr. Florian Putz}
\def\corrEmail{florian.putz@uk-erlangen.de}

\begin{document}
\onecolumn

\maketitle

\begin{abstract}
    \textbf{Background and Purpose:} Manual delineation of target volumes in head and neck cancer (HNC) remains a significant bottleneck in radiotherapy planning, characterized by high inter-observer variability and time consumption. This study evaluates the integration of a Volume-Aware (VA) Dice loss function into a self-configuring deep learning framework to enhance the auto-segmentation of primary tumors (PT) and metastatic lymph nodes (LN) for adaptive MR-guided radiotherapy. We specifically investigate how volume-sensitive weighting affects the detection of small nodal metastases compared to conventional loss functions.\\
    \textbf{Materials and Methods:} Utilizing the HNTS-MRG 2024 challenge dataset, we implemented an nnU-Net ResEnc M architecture. We conducted a multi-label segmentation task, comparing a standard Dice loss baseline against two Volume-Aware configurations: a "Dual Mask" setup (VA loss on both PT and LN) and a "Selective LN Mask" setup (VA loss on LN only). Evaluation metrics included volumetric Dice scores, surface-based metrics (SDS, MSD, HD95), and lesion-wise binary detection sensitivity and precision.\\
    \textbf{Results:} The Selective LN Mask configuration achieved the highest LN Volumetric Dice Score ($0.758$ vs. $0.734$ baseline) and significantly improved LN Lesion-Wise Detection Sensitivity ($84.93\%$ vs. $81.80\%$). However, a critical trade-off was observed; PT detection precision declined significantly in the selective setup ($63.65\%$ vs. $81.27\%$). The Dual Mask configuration provided the most balanced performance across both targets, maintaining primary tumor precision at $82.04\%$ while improving LN sensitivity to $83.46\%$.\\
    \textbf{Conclusions:} A volume-sensitive loss function mitigated the under-representation of small metastatic lesions in HNC. While selective weighting yielded the best nodal detection, a dual-mask approach was required in a multi-label segmentation task to maintain geometric precision for larger primary tumor volumes.
\end{abstract}

\section{Introduction}
Head and neck cancer (HNC) represents a major global health burden, accounting for approximately 4.8\% of all cancer incidences and 4.7\% of cancer-related fatalities annually \citep{sung2021global, argiris2008head}. Radiotherapy (RT) is a fundamental pillar of HNC management, frequently employed as a definitive or adjuvant treatment modality. The success of RT is predicated on the precise delineation of target volumes, specifically the primary gross tumor volume (GTVp) and metastatic lymph nodes (GTVn), to ensure tumor control while minimizing dose to adjacent organs-at-risk (OARs) \citep{kosmin2019rapid}.
\begin{figure*}[htbp]
    \centering
        \includegraphics[width=\textwidth]{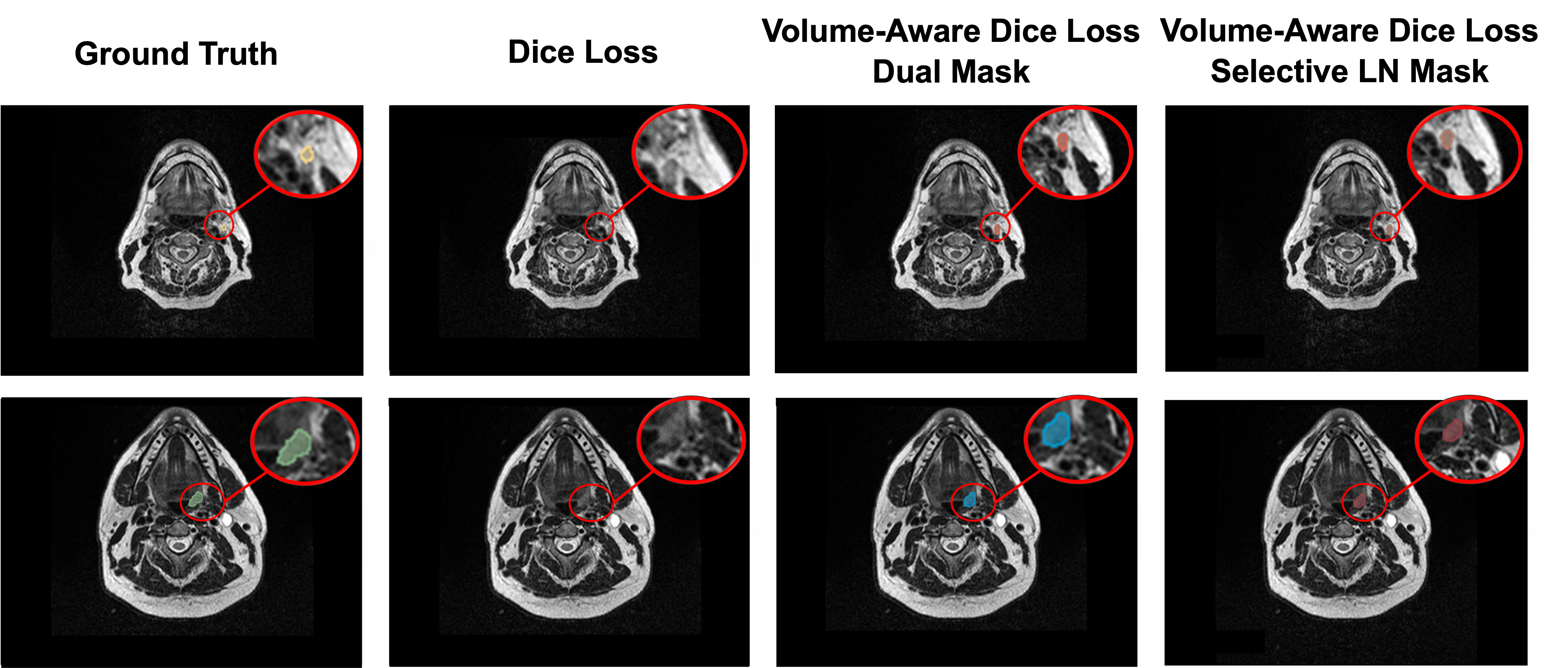}

    \caption{Examples of better performance of the VA dice loss to the baseline}
    \label{fig:example_image_1}
\end{figure*}
The manual contouring process is notoriously time-intensive, often requiring up to three hours per case for an experienced radiation oncologist \citep{njeh2008tumor, kihara2023clinical, jin2022towards}. Moreover, HNC is characterized by complex anatomy and significant inter-observer variability (IOV), which can lead to inconsistencies in treatment delivery \citep{brouwer20123d}. In modern radiotherapy, the transition toward personalized medicine has been accelerated by the introduction of Magnetic Resonance (MR)-guided adaptive radiotherapy (ART). MR-guided ART offers superior soft-tissue contrast compared to conventional Computed Tomography (CT), allowing for continuous anatomical adaptation to account for shifts in anatomy, tumor shrinkage, or patient weight loss during the multi-week treatment course \citep{aristophanous2024clinical, outeiral2021oropharyngeal}.

However, for ART to be clinically viable on a daily basis, the segmentation process must be automated, rapid, and highly reliable. Manual contouring on a daily basis is unfeasible in high-volume clinics. Consequently, the development of robust auto-segmentation tools is a prerequisite for the widespread adoption of ART.

In recent years, deep learning-based auto-segmentation, particularly the U-Net architecture and its self-configuring framework, nnU-Net, has set new benchmarks in medical imaging \citep{ronneberger2015u, isensee2021nnu}. Despite these advancements, a persistent issue in HNC segmentation is the "volume imbalance" between the typically large primary tumor and the numerous, variably sized, and often very small lymph node metastases. Standard loss functions, such as the soft Dice loss, are dominated by larger structures, which leads to under-representation of the smaller lesions \citep{chartrand2022automated, hu2019multimodal, huang2022deep}.

This research investigates the clinical implementation of a Volume-Aware Dice loss for HNC. Unlike conventional approaches, this loss function assigns higher weights to the voxels of smaller target volumes, effectively normalizing the influence of disparate target sizes. We examine two distinct multi-label strategies, "Dual Mask" and "Selective LN Mask", to determine how targeted volume-weighting influences the sensitive balance between primary tumor and nodal target segmentation accuracy.

\section{Materials and Methods}
\subsection{Dataset Description}
The study utilized the official dataset of the Head and Neck Tumor Segmentation for MR-Guided Applications (HNTS-MRG) 2024 challenge \citep{hntsmrg2024}. This dataset represents one of the most comprehensive longitudinal collections of HNC MRI data, comprising scans from patients with histologically confirmed oropharyngeal or hypopharyngeal carcinomas. The target structures were annotated on T2-weighted (T2w) MRI scans including both fat-suppressed and non-fat-suppressed acquisitions:
\begin{itemize}
    \item \textbf{PT (Primary Tumor):} The GTVp site, typically characterized by larger, continuous volumes.
    \item \textbf{LN (Lymph Nodes):} The GTVn, including all metastatic regional nodes, often presenting as small, fragmented, and anatomically dispersed structures.
\end{itemize}

To ensure the highest quality of ground truth, annotations were established via the STAPLE (Simultaneous Truth and Performance Level Estimation) algorithm, reflecting a consensus from multiple expert observers and reviewed by senior radiation oncologists \citep{hntsmrg2024}. The publicly available training partition of 150 cases was used in this study, evaluated via 5-fold cross-validation with three repeats \citep{hntsmrg2024}.

\subsection{nnU-Net ResEnc Framework}
We employed the nnU-Net framework, which serves as a state-of-the-art benchmark for medical image segmentation due to its ability to self-configure its entire pipeline based on a dataset's characteristics ("fingerprint") \citep{isensee2021nnu}. Specifically, we utilized the "ResEnc M" configuration. 

The ResEnc M variant incorporates residual blocks into the encoder path of the U-Net. Residual connections facilitate the training of deeper architectures by allowing gradients to flow through skip-connections. 

Beyond the architecture, nnU-Net automates preprocessing steps, such as voxel spacing normalization to a median resolution, z-score intensity normalization based on the foreground voxels, and extensive data  augmentation (rotations, scaling, mirroring, and intensity perturbations) to improve the model's robustness against the variability of the included MRI scans.

\subsection{Volume-Aware (VA) Dice Loss}
The standard Dice similarity coefficient (DSC) is defined as the overlap between prediction and ground truth. In clinical HNC datasets, the primary tumor can be several orders of magnitude larger than a single metastatic lymph node. In addition, individual lymph node metastases vary largely in size. In the context of a standard Dice loss:
\begin{equation}
L_{Dice} = 1 - \frac{2 \sum p_i g_i + \epsilon}{\sum p_i^2 + \sum g_i^2 + \epsilon}
\end{equation}
\begin{figure}[]
    \centering
    \includegraphics[width=0.48\textwidth]{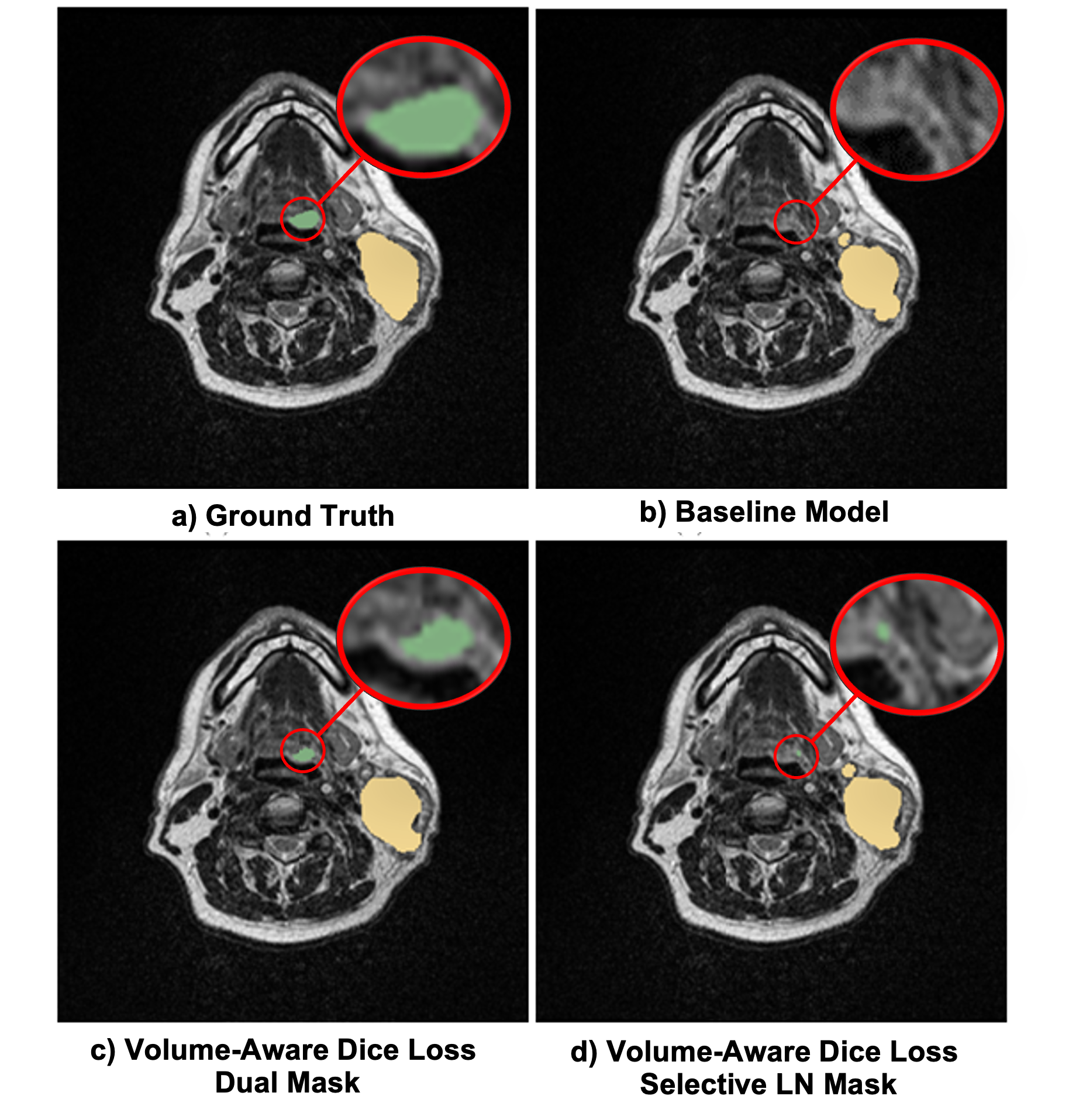}
    \caption{Segmentation masks produced using different configurations on the multi-label
dataset. The visualization illustrates the influence of the Volume-Aware Dice loss under
various settings. (a) Ground Truth: PT is shown in green, and LN in yellow. (b) Baseline
Model: accurate segmentation for LN, but underperformance in PT segmentation. (c)
Volume-Aware Dice Loss - Dual Mask Configuration: segmentation quality for both PT
(green) and LN (yellow) is well maintained. (d) Volume-Aware Dice Loss - Selective LN
Mask Configuration: achieves improved LN segmentation compared to the dual mask setup,
while PT performance decreases slightly, yet remains better than in the base model.}
    \label{fig:example_image_2}
\end{figure}
This often leads to models that "ignore" smaller nodal structures in favor of perfecting the boundaries of the larger primary site.

To mitigate this, we integrated a Volume-Aware (VA) Dice loss, also referred to as an Adaptive Dice loss \citep{hu2019multimodal, chartrand2022automated}. The loss function $L_{vol-dice}$ incorporates a diagonal weighting matrix $W$ that assigns a weight $w_i$ to each voxel $i$ based on the volume of the connected component to which it belongs:
\begin{equation}
L_{vol-dice} = - \frac{Cg^{T}Wp + \epsilon}{p^{T}p + g^{T}Wg + \epsilon}
\end{equation}
Specifically, $w_i = 1/\sqrt{V_j}$ for a voxel belonging to a lesion of volume $V_j$. By using the inverse square root of the volume, the loss ensures that smaller lesions contribute more significantly to the total loss. This mechanism increases the relative penalty for misclassifying voxels belonging to small, clinically critical nodal metastases during training, reducing the risk that the model converges toward solutions that overlook them in favor of larger structures.

\subsection{Experimental Configurations}
To investigate the interplay between large and small targets in a multi-label auto-segmentation setting, three strategies were compared:
\begin{enumerate}
    \item \textbf{Baseline:} Default nnU-Net trained with a combined Dice and Cross-Entropy loss.
    \item \textbf{Dual Mask VA:} VA Dice loss applied to both PT and LN masks simultaneously, aiming for volume-sensitive delineation across all targets.
    \item \textbf{Selective LN Mask VA:} VA loss applied exclusively to the LN class, while the PT channel retained a standard Dice loss. This configuration tests whether prioritizing nodal sensitivity can be achieved without affecting the primary tumor's representation.
\end{enumerate}

\subsection{Evaluation Metrics and Statistical Analysis}
Segmentation performance was assessed using an automated evaluation pipeline computing volumetric, surface-based, and lesion-level metrics. Volumetric Dice Scores quantified overall overlap between predicted and ground truth masks. Surface Dice Score (SDS, tolerance \SI{1}{\milli\meter}) and Mean Surface Distance (MSD) evaluated boundary alignment, while the 95th percentile Hausdorff Distance (HD95) captured worst-case boundary deviations. Surface distances were computed using the \texttt{surface-distance} Python library \citep{surface_distance}. 

For lesion-level evaluation, binary detection metrics were calculated by treating each connected component as an individual detection unit. A ground truth lesion was counted as detected (true positive for sensitivity) if it overlapped with any predicted lesion, and vice versa for precision. This approach captures the model's ability to detect small, clinically relevant nodal metastases that may have minimal impact on aggregate volumetric metrics.

All experiments were conducted using 5-fold cross-validation with three independent training repetitions per fold to account for stochastic variation in weight initialization and training dynamics. Final metrics were computed as the mean across all repetitions and folds. Statistical significance between configurations was assessed using the Wilcoxon signed-rank test applied to paired per-patient scores, with $p < 0.05$ considered statistically significant.

\section{Results}

\subsection{Volumetric and Surface Similarity}

\begin{figure*}[htbp]
    \centering
        \includegraphics[width=\textwidth]{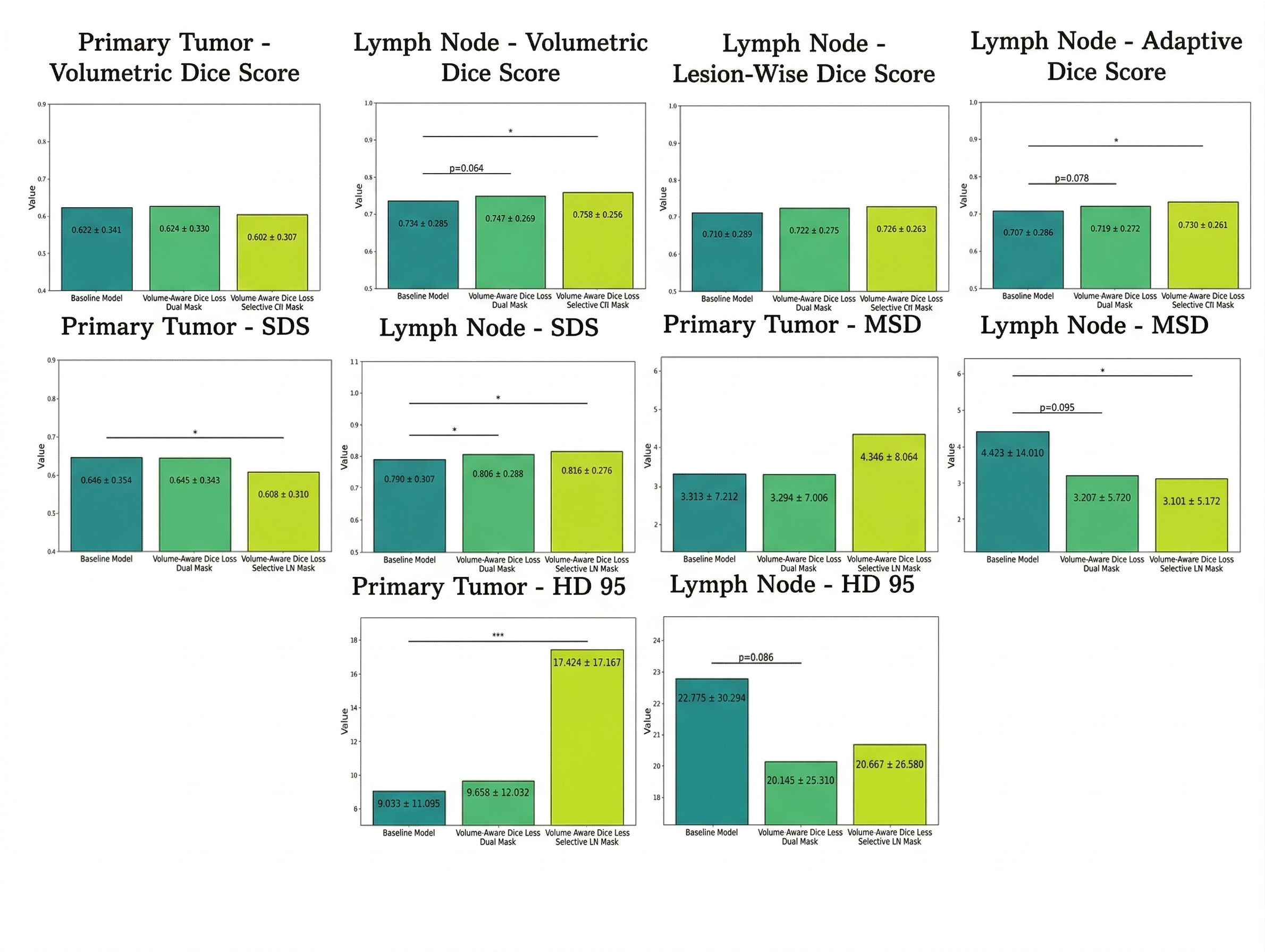}

    \caption{Plots showing volumetric and surface distance similarities for primary tumor and lymph node metastases across evaluated configurations.}
    \label{fig:example_image_3}
\end{figure*}
As it is shown in Fig. 3 and Table 1, the integration of the VA Dice loss led to a consistent and measurable improvement in lymph node (LN) segmentation accuracy. The Selective LN configuration achieved the highest LN Dice score of 0.758, a significant increase from the 0.734 achieved by the baseline model. Beyond volumetric overlap, distance-based and surface-based metrics showed substantial gains in terms of boundary alignment for LN targets.

The Dual Mask configuration also outperformed the baseline for LN targets, yielding a Dice score of 0.747 and maintaining a more stable performance for the PT (Dice 0.624 vs. 0.622 baseline).

\begin{table*}[htbp]
\centering
\caption{Comparison of segmentation performance across models.}
\label{tab:segmentation_results}
\resizebox{\linewidth}{!}{%

\begin{tabular}{
l
S[table-format=1.3]
S[table-format=1.3]
S[table-format=1.3]
S[table-format=2.3]
S[table-format=1.3]
}
\toprule
\textbf{Metric} &
\textbf{Baseline Model} &
\textbf{Volume-Aware Dice Loss Dual Mask} &
\textbf{Dual Mask p value} &
\textbf{Volume-Aware Dice Loss Selective LN Mask} &
\textbf{Selective LN Mask p value} \\
\midrule
PT Dice Score               & 0.622 & 0.624 & 0.949 & 0.602  & 0.184 \\
LN Dice Score               & 0.734 & 0.747 & 0.064 & 0.758  & 0.019 \\
LN Adaptive Dice Score     & 0.707 & 0.719 & 0.078 & 0.730  & 0.020 \\
LN Lesion-Wise Dice Score  & 0.710 & 0.722 & 0.235 & 0.726  & 0.224 \\
PT SDS (mm)                     & 0.646 & 0.645 & 0.752 & 0.608  & 0.018 \\
LN SDS (mm)                     & 0.790 & 0.806 & 0.040 & 0.816  & 0.014 \\
PT MSD (mm)                      & 3.313 & 3.294 & 0.797 & 4.346  & 0.351 \\
LN MSD (mm)                      & 4.423 & 3.207 & 0.095 & 3.101  & 0.047 \\
PT HD 95 (mm)                    & 9.033 & 9.658 & 0.124 & 17.424 & 0.000 \\
LN HD 95 (mm)                    & 22.775 & 20.145 & 0.086 & 20.667 & 0.453 \\
\bottomrule
\end{tabular}%
}
\end{table*}

\subsection{Detection Sensitivity and Precision}

\begin{table*}[htbp]
\centering
\caption{Lesion-wise binary detection precision and sensitivity compared across models.}
\label{tab:binary_metrics}
\resizebox{\linewidth}{!}{%
\begin{tabular}{lccccc}
\toprule
 &
\textbf{Baseline Model} &
\textbf{Volume-Aware Dice Loss Dual Mask} &
\textbf{Dual Mask  p value} &
\textbf{Volume-Aware Dice Loss Selective LN Mask} &
\textbf{Selective LN Mask p value} \\
\midrule
PT Lesion-Wise Binary Precision      & 0.813 & 0.820 & 0.510 & 0.636 & 0.000 \\
LN Lesion-Wise Binary Precision      & 0.736 & 0.736 & 0.983 & 0.760 & 0.104 \\
PT Lesion-Wise Binary Sensitivity    & 0.837 & 0.848 & 0.298 & 0.870 & 0.038 \\
LN Lesion-Wise Binary Sensitivity    & 0.818 & 0.835 & 0.078 & 0.849 & 0.010 \\
\bottomrule
\end{tabular}%
}
\end{table*}
Lesion-wise evaluation using binary detection metrics (Table 2 and Fig. 4) confirmed that VA loss fundamentally altered the model's detection capabilities. The Selective LN configuration reached a nodal sensitivity of 84.93\%, identifying lymphatic nodules that the baseline model failed to detect. However, this increased sensitivity for small structures came at the cost of reduction in Precision for the primary tumor.

In the Selective LN configuration, PT precision dropped to 63.65\%, a significant decline from the baseline's 81.27\%. This suggests that the network's focus on small nodal structures led to an increase in false positives or fragmented segmentations in the larger primary tumor region.
\begin{figure*}[htbp]
    \centering
        \includegraphics[width=\textwidth]{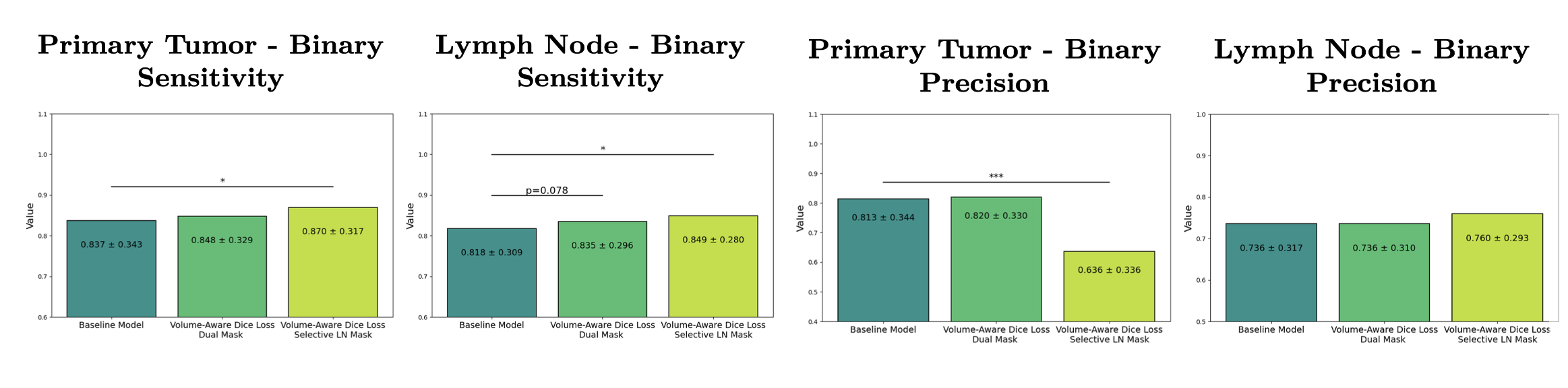}

    \caption{Plots showing binary detection scores (Sensitivity and Precision) for primary tumor and lymph node metastasis across evaluated configurations.}
    \label{fig:example_image_4}
\end{figure*}

\section{Discussion}
The results of this study highlight a fundamental "precision-sensitivity trade-off" in multi-label HNC auto-segmentation. While Volume-Aware Dice loss successfully addresses small lesion detection, a critical requirement for selective nodal irradiation, it creates a competitive gradient environment between structures of disparate sizes. When the network is heavily penalized for missing small nodes, it may become "hyper-sensitive," leading to over-segmentation of normal tissue structures as malignant nodal tissue and decreasing segmentation accuracy for larger targets like the primary tumor.

The phenomenon of this sharp drop in PT precision in the selective setup, is a key finding. It suggests that selective weighting of the lymph node class can impair the feature representation for the larger, continuous primary tumors. In clinical terms, while a high sensitivity is desired to ensure all tumor tissue is included in the high-dose volume, low precision in the primary tumor region could lead to unnecessary irradiation of healthy tissues, particularly organs-at-risk located in the immediate tumor periphery.

The Dual Mask VA Dice loss strategy could provide a more balanced resolution to this trade-off. By applying VA loss to both target classes, the network is better optimized for segmentation of all target types while still preventing the under-representation of smaller lesions. This strategy maintained a high precision for the PT (82.04\%) while still improving LN sensitivity (83.46\%) compared to the baseline (Fig. 4). 

Furthermore, including the PT label in a multi-label segmentation task provides additional supervisory signal that may improve overall segmentation accuracy. Primary tumors and their derived lymph node metastases share imaging characteristics on T2-weighted MRI and are anatomically connected through predictable lymphatic drainage pathways, often in close spatial proximity. Joint segmentation of both structures allows the network to exploit these shared features and spatial relationships, mirroring clinical practice where PT and LN are typically delineated as part of the same contouring task.

\begin{figure}[htbp]
    \centering
        
        \includegraphics[width=0.48\textwidth]{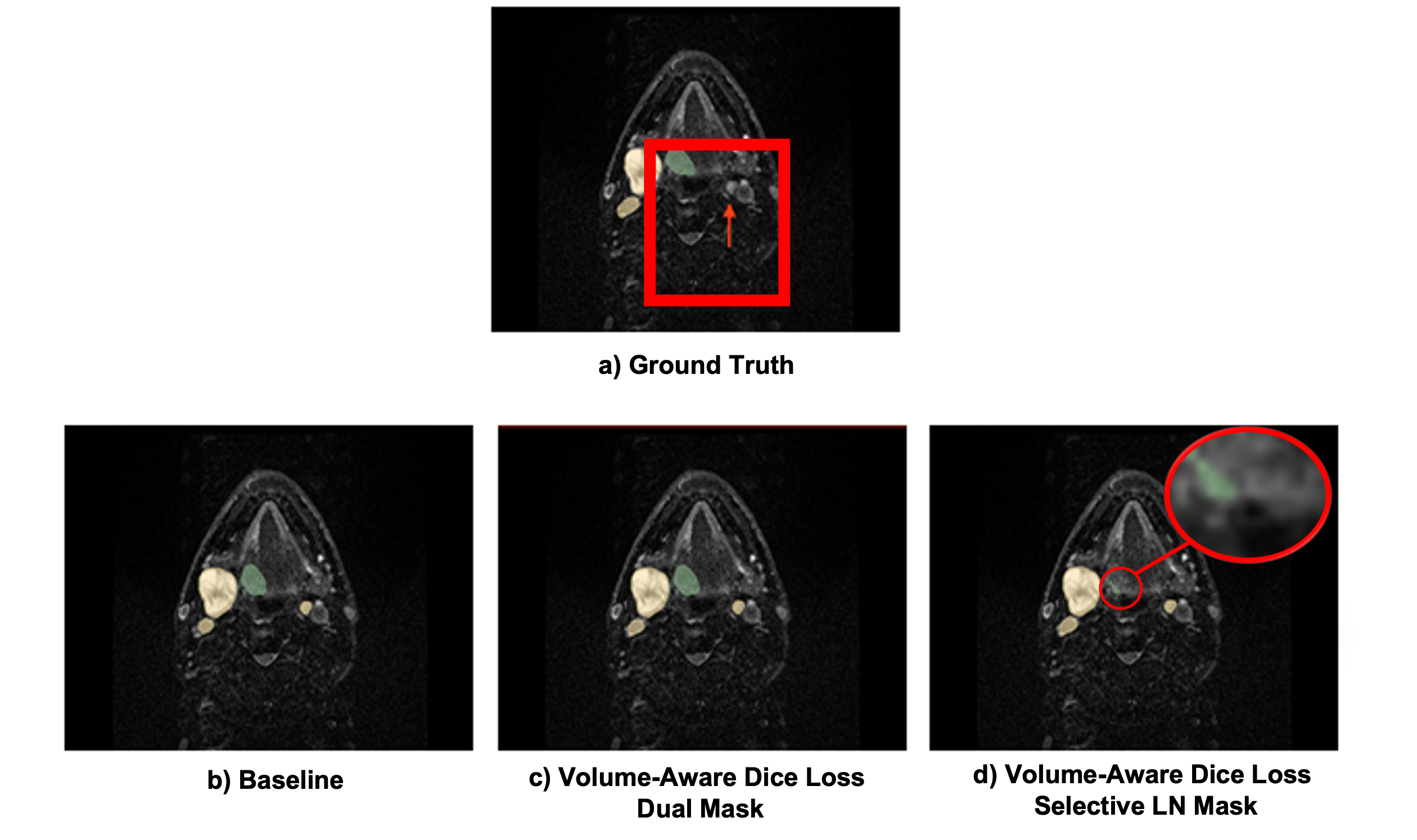}
    
    \caption{A comprehensive visualization showcasing all evaluated configurations on a
single patient. (a) Ground Truth: PT is displayed in green, and LN in yellow. (b, c, d): (c) achieving the best overall performance, exceeding both the base model (b) and the Volume-Aware Dice loss in selective LN mask
approach (d). Arrow in ground truth mask: False positive lymph node metastasis corresponding to an anatomical structure with suspicious imaging appearance.}
    \label{fig:example_image_5}
\end{figure}

\section{Conclusion}
This study demonstrates that the integration of customized Volume-Aware Dice loss function into the state-of-the-art nnU-Net architecture could significantly improve the auto-segmentation of radiotherapy targets in HNC. Prioritizing the detection of small, variable nodal metastases, could achieve higher clinical precision for radiotherapy planning. While selective weighting yielded the best nodal segmentation results, a dual-mask approach is recommended in clinical multi-label tasks to avoid sacrificing precision in larger primary targets. These findings could ultimately help in more efficient and precise MR-guided adaptive radiotherapy, reducing the manual burden on clinical teams and potentially improving patient outcomes through consistent and sensitive target delineation.

\end{document}